\documentclass{llncs}
\usepackage{amsmath}
\usepackage{amssymb}
\usepackage{amsfonts}
\usepackage{makeidx}

\input{tcilatex}
\begin{document}

\frontmatter
\author{Andreas Maurer\inst{1} and Massimiliano Pontil\inst{2}}

\mainmatter

\title{$K$-Dimensional Coding Schemes in Hilbert Spaces}

\titlerunning{Transformation Classes} 

\authorrunning{A. Maurer and M. Pontil}

\institute{Adalbertstrasse 55\\D-80799 M\"unchen, Germany\\am@andreas-maurer.eu
\and 
Dept. of Computer Science\\University College London\\Malet Pl., WC1E, 
London, UK\\m.pontil@cs.ucl.ac.uk
}

\maketitle

\begin{abstract}
This paper presents a general coding method where data in a Hilbert
space are represented by finite dimensional coding vectors. The method
is based on empirical risk minimization within a certain class of
linear operators, which map the set of coding vectors to the Hilbert
space. Two results bounding the expected reconstruction error of the
method are derived, which highlight the role played by the codebook
and the class of linear operators. The results are specialized to some
cases of practical importance, including $K$-means clustering,
nonnegative matrix factorization and other sparse coding methods.
\end{abstract}


\titlerunning{Transformation Classes}

\authorrunning{A. Maurer and M. Pontil}

\section{Introduction}

We study a general class of $K$-dimensional coding methods for data drawn
from a distribution $\mu $ on the unit ball of a Hilbert space $H$. These
methods encode a data point $x\sim \mu $ as a vector $\hat{y}\in \mathbb{R}^{K}$, according to the formula 
\begin{equation*}
\hat{y}=\arg \min_{y\in Y}\left\Vert x-Ty\right\Vert ^{2},
\end{equation*}
where $Y\subseteq \mathbb{R}^{K}$ is a prescribed set of \textit{codes}
(called the \emph{codebook}), which we can always assume to span $\mathbb{R}^{K}$, and $T:\mathbb{R}^{K}\rightarrow H$ is a linear map, which defines a
particular \textit{implementation} of the codebook. It embeds the codebook $Y
$ in $H$ and yields the set $T\left( Y\right) $ of exactly codable patterns.
If $\hat{y}$ is the code found for $x$ then $\hat{x}=T\hat{y}$ is the
reconstructed data point. The quantity 
\begin{equation*}
f_{T}\left( x\right) =\min_{y\in Y}\left\Vert x-Ty\right\Vert ^{2}
\end{equation*}%
is called the reconstruction error.

Given a codebook $Y$ and a finite number of independent observations $%
x_{1},\dots ,x_{m}\sim \mu $, a common sense approach searches for an
implementation ${\hat T}$ which is optimal on average over the
observed points, that is 
\begin{equation}
{\hat T}=\arg \min_{T\in \mathcal{T}}\frac{1}{m}\sum_{i=1}^{m}f_{T}%
\left( x_{i}\right) ,  \label{main algorithm}
\end{equation}%
where $\mathcal{T}$ denotes some class of linear maps $T:\mathbb{R}%
^{K}\rightarrow H$. As we shall see, this framework is general enough to
include principal component analysis, $K$-means clustering, non-negative
matrix factorization \cite{Lee 1999} and the sparse coding method as
proposed in \cite{Olshausen 1996}.

Whenever the codebook $Y$ is compact and $\mathcal{T}$ is bounded in the
operator norm this approach is justified by the following high-probability,
uniform bound on the expected reconstruction error.

\begin{theorem}
\label{Theorem general} Suppose that $Y$ is a closed subset of the unit ball
of $\mathbb{R}^{K}$, that there is $c\geq 1$ such that $\left\Vert
T\right\Vert _{\infty }\leq c$ for all $T\in \mathcal{T}$ and that $\delta
\in \left( 0,1\right) $. Then with probability at least $1-\delta $ in the
observed data $x_1,\dots,x_m \sim \mu$ we have for every $T\in \mathcal{T}$
that 
\begin{equation*}
\mathbb{E}_{x\sim \mu }f_{T}\left( x\right) -\frac{1}{m}\sum_{i=1}^{m}f_{T}%
\left( x_{i}\right) \leq 6c^{2}K^{2}\sqrt{\frac{\pi }{m}}+c^2\sqrt{\frac{%
8\ln 1/\delta }{m}}.
\end{equation*}%
The bound is two-sided in the sense that also with probability at least $%
1-\delta $ we have for every $T\in \mathcal{T}$ that%
\begin{equation*}
\frac{1}{m}\sum_{i=1}^{m}f_{T}\left( x_{i}\right) -\mathbb{E}_{x\sim \mu
}f_{T}\left( x\right) \leq 6c^{2}K^{2}\sqrt{\frac{\pi }{m}}+c^{2}\sqrt{\frac{%
8\ln 1/\delta }{m}}.
\end{equation*}
\end{theorem}

Any compact subset of $\mathbb{R}^{K}$ can of course be down-scaled to be
contained in the unit ball, and the scaling factor can be absorbed in $c$,
so that the above result is applicable to any compact codebook.

The theorem implies a bound on the excess risk: let $T_{0}\in
\mathcal{T} $ be a minimizer of the expected reconstruction error 
within the set $\mathcal{T}$. It
follows from the definition of ${\hat T}$ and the above result that
the expected reconstruction error of ${\hat T}$ is with high
probability not more than $O\left( 1/\sqrt{m}\right) $ worse than
that of $T_{0}$.

This order in $m$ is optimal, as we know from existing lower bounds for $K$%
-means clustering \cite{Bartlett Lugosi 1998}. The above dependence on $K$
is, however, generally not optimal, and can be considerably improved with a
more careful analysis, if we are prepared to accept the slightly inferior
rate of $\sqrt{\ln m/m}$ in the sample size. To state this improvement define%
\begin{equation*}
\left\Vert \mathcal{T}\right\Vert _{Y}=\sup_{T\in \mathcal{T}}\left\Vert
T\right\Vert _{Y}=\sup_{T\in \mathcal{T}}\sup_{y\in Y}\left\Vert
Ty\right\Vert .
\end{equation*}%
We then have the following result.

\begin{theorem}
\label{Theorem main} Assume that $\left\Vert \mathcal{T}\right\Vert _{Y}\geq
1$ and that the functions $f_{T}$ for $T\in \mathcal{T}$, when restricted to
the unit ball of $H$, have range contained in $\left[ 0,b\right] $. Fix $%
\delta >0$.

Then with probability at least $1-\delta $ in the observed data $x_1,\dots,x_m \sim \mu$ we have for every $T\in \mathcal{T}$ that 
\begin{equation*}
\mathbb{E}_{x\sim \mu }f_{T}\left( x\right) -\frac{1}{m}\sum_{i=1}^{m}f_{T}%
\left( x_{i}\right) \leq \frac{K}{\sqrt{m}}\left( 14\left\Vert \mathcal{T}%
\right\Vert _{Y}+\frac{b}{2}\sqrt{\ln \left( 16m\left\Vert \mathcal{T}%
\right\Vert _{Y}^{2}\right) }\right) +b\sqrt{\frac{\ln 1/\delta }{2m}}.
\end{equation*}%
The bound is two sided in the same sense as the previous result.
\end{theorem}

Both results immediately imply uniform convergence in probability. We are
not aware of other results for nonnegative matrix factorization \cite{Lee
1999} or the sparse coding techniques as in \cite{Olshausen 1996}.

Before proving our results, we will illustrate their implications in some cases
of interest. It turns out that the dependence on $K$ in Theorem \ref{Theorem
main} adapts to the specific situation under consideration.

A preliminary version of this paper appeared in the proceedings of the 
2008 Algorithmic Learning Theory Conference \cite{MauPon}. The new version contains Theorem
\ref{Theorem general} and a simplified proof of 
Theorem \ref{Theorem main} with improved constants.

\section{Examples of coding schemes\label{section examples}}

Several coding schemes can be expressed in our framework. We describe some of these methods and how our result applies.

\subsection{Principal component analysis\label{subsection pca}}

Principal component analysis (PCA) seeks a $K$-dimensional orthogonal
projection which maximizes the projected variance and then uses this
projection to encode future data. A projection $P$ can be expressed as $%
TT^{\ast }$ where $T$ is an isometry which maps $\mathbb{R}^{K}$ to the
range of $P$. Since 
\begin{equation*}
\left\Vert Px\right\Vert ^{2}=\left\Vert x\right\Vert ^{2}-\Vert x-Px\Vert
^{2}=\Vert x\Vert ^{2}-\min_{y\in \mathbb{R}^{K}}\left\Vert x-Ty\right\Vert
^{2}
\end{equation*}%
finding $P$ to maximize the true or empirical expectation of $\Vert
Px\Vert ^{2}$ is equivalent to finding $T$ to minimize the
corresponding expectation of $\min_{y\in \mathbb{R}^{K}}\left\Vert
x-Ty\right\Vert ^{2}$. We see that PCA is described by our framework
upon the identifications $Y=
\mathbb{R}^{K}$ and $\mathcal{T}$ is restricted to the class of isometries $%
T:\mathbb{R}^{K}\rightarrow H$. Given $T\in \mathcal{T}$ and $x\in H$ the
reconstruction error is 
\begin{equation*}
f_{T}\left( x\right) =\min_{y\in \mathbb{R}^{K}}\left\Vert x-Ty\right\Vert
^{2}.
\end{equation*}%
If the data are constrained to be in the unit ball of $H$, as we generally
assume, then it is easily seen that we can take $Y$ to be the unit ball of $%
\mathbb{R}^{K}$ without changing any of the encodings. We can therefore
apply Theorem \ref{Theorem main} with $\left\Vert \mathcal{T}\right\Vert _{Y}=1$ and $b=1$.
This is besides the point however, because in the simple case of PCA much
better bounds are available (see \cite{ShaweTaylor 2005}, \cite{Zwald} and
Lemma \ref{Lemma PCA} below). In \cite{Zwald} local Rademacher averages are
used to give faster rates under certain circumstances.

An objection to PCA is, that generic codes have $K$ nonzero components,
while for practical and theoretical reasons sparse codes with much less than 
$K$ nonzero components may be preferable \cite{Olshausen 1996}.

\subsection{$K$-means clustering or vector quantization\label{subsection
kmeans clustering}}

Here $Y=\left\{ e_{1},\dots ,e_{K}\right\} $, where the vectors $e_{k}$ form an
orthonormal basis of $\mathbb{R}^{K}$. An implementation $T$ now defines a
set of centers $\left\{ Te_{1},\dots ,Te_{K}\right\} $, the reconstruction
error is $\min_{k=1}^{K}\left\Vert x-Te_{k}\right\Vert ^{2}$ and a data
point $x$ is coded by the $e_{k}$ such that $Te_{k}$ is nearest to $x$. The
algorithm (\ref{main algorithm}) becomes 
\begin{equation*}
{\hat T}=\arg \min_{T\in \mathcal{T}}\frac{1}{m}\sum_{i=1}^{m}%
\min_{k=1}^{K}\left\Vert x_{i}-Te_{k}\right\Vert ^{2}.
\end{equation*}%
It is clear that every center $Te_{k}$ has at most unit norm, so that $%
\left\Vert \mathcal{T}\right\Vert _{Y}=1$. Since all data points are in the
unit ball we have $\left\Vert x-Te_{k}\right\Vert ^{2}\leq 4$ so we can set $%
b=4$ and the bound in Theorem \ref{Theorem main} becomes 
\begin{equation*}
\left( 14+2\sqrt{\ln \left( 16m\right) }\right) \frac{K}{\sqrt{m}}+\sqrt{%
\frac{8\ln \left( 1/\delta \right) }{m}}.
\end{equation*}

The order of this bound matches up to $\sqrt{\ln m}$ the order given in \cite%
{Biau Devroye Logosi 2006} or \cite{ShaweTaylor 2007}. To illustrate our
method we will also prove the bound 
\begin{equation*}
\sqrt{18\pi }\frac{K}{\sqrt{m}}+\sqrt{\frac{8\ln \left( 1/\delta \right) }{m}%
}
\end{equation*}%
(Theorem \ref{Theorem Kmeans clustering}), which is essentially the same as
those in \cite{Biau Devroye Logosi 2006} or \cite{ShaweTaylor 2007}. There
is a lower bound of order $\sqrt{K/m}$ in \cite{Bartlett Lugosi 1998}, and
it is unknown which of the two bounds (upper or lower) is tight.

In $K$-means clustering every code has only one nonzero component, so that
sparsity is enforced in a maximal way. On the other hand this results in a
weaker approximation capability of the coding scheme.

\subsection{Nonnegative matrix factorization\label{subsection nonnegative
matrix factorization}}

Here $Y$ is the positive orthant in $\mathbb{R}^K$, that is the cone 
\begin{equation*}
Y=\left\{y: y=(y_1,\dots,y_K),~y_k \geq 0, 1 \leq k \leq K \right\}.
\end{equation*}
A chosen map $T$ generates a cone $T\left( Y\right) \subset H$ onto which
incoming data is projected. In the original formulation by Lee and Seung 
\cite{Lee 1999} it is postulated that both the data and the vectors $Te_{k}$
be contained in the positive orthant of some finite dimensional space, but
we can drop most of these restrictions, keeping only the requirement that $%
\left\langle Te_{k},Te_{l}\right\rangle \geq 0$ for $1\leq k,l\leq K$.

No coding will change if we require that $\left\Vert Te_{k}\right\Vert =1$
for all $1\leq k\leq K$ by a suitable normalization. The set $\mathcal{T}$
is then given by 
\begin{equation*}
\mathcal{T}=\{T:T\in \mathcal{L}(\mathbb{R}^{K},H),~\left\Vert
Te_{k}\right\Vert =1,~\left\langle Te_{k},Te_{l}\right\rangle \geq 0,~1\leq
k,l\leq K\}.
\end{equation*}%
We can restrict $Y$ to its intersection with the unit ball in $\mathbb{R}^{K}
$ (see Lemma \ref{Little Lemma} below). We obtain that $\left\Vert \mathcal{T
}\right\Vert _{Y}=\sqrt{K}$. Hence, Theorem \ref{Theorem main} yields the
bound 
\begin{equation*}
\frac{K}{\sqrt{m}}\left( 14\sqrt{K}+\frac{1}{2}\sqrt{\ln \left( 16mK\right) }%
\right) +\sqrt{\frac{\ln \left( 1/\delta \right) }{2m}}
\end{equation*}%
on the estimation error. We do not know of any other generalization bounds
for this coding scheme.

Nonnegative matrix factorization appears to encourage sparsity, but cases
have been reported where sparsity was not observed \cite{Li 2001}. In fact
this undesirable behavior should be generic for exactly codable data.
Various authors have therefore proposed additional constraints (\cite{Li
2001}, \cite{Hoyer 2004}). It is clear that additional constraints on $%
\mathcal{T}$ can only improve estimation and that the passage from $Y$ to a
subset can only improve our bounds, because the quantity $\|\mathcal{T}\|_Y$
would decrease.

\subsection{Sparse coding\label{subsection sparse coding}}

Another method arises by choosing the $\ell _{p}$-unit ball as a codebook.
Let $Y=\{y:y\in \mathbb{R}^{K},~\Vert y\Vert _{p}\leq 1\}$ and $\mathcal{T}
=\{T:\mathbb{R}^{K}\rightarrow H:\Vert Te_{k}\Vert \leq 1,1\leq k\leq K\}$.
We have 
\begin{equation*}
\Vert Ty\Vert =\Vert \sum_{k=1}^{K}y_{k}Te_{k}\Vert \leq
\sum_{k=1}^{K}|y_{k}|\Vert Te_{k}\Vert \leq \left( \sum_{k=1}^{K}\Vert
Te_{k}\Vert ^{q}\right) ^{1/q}\leq K^{1/q}=K^{1-1/p}
\end{equation*}%
implying that $\Vert \mathcal{T}\Vert _{Y}\leq K^{1-1/p}$.

By the same argument as above all the $f_{T}$ have range contained in $\left[
0,1\right] $, so Theorem \ref{Theorem main} can be applied with $b=1$ to
yield the bound 
\begin{equation*}
\frac{K}{\sqrt{m}}\left( 14K^{1-1/p}+\frac{1}{2}\sqrt{\ln \left(
16mK^{2-2/p}\right) }\right) +\sqrt{\frac{\ln \left( 1/\delta \right) }{2m}}
\end{equation*}%
on the estimation error. The best bound is obtained when $p=1$, and the
order in $K$ matches that of the bound for $K$-means clustering described
earlier.

The method for $p=1$ is similar to the sparse-coding method proposed by
Olshausen and Field \cite{Olshausen 1996}, with the difference that the term 
$\Vert y\Vert _{1}$ is used as a penalty term instead of the hard constraint 
$\Vert y\Vert _{1}\leq 1$. The method of Olshausen and Field \cite{Olshausen
1996} approximates with a compromise of geometric proximity and sparsity and
our result asserts that the observed value of this compromise generalizes to
unseen data if enough data have been observed.

\section{Proofs\label{section proofs}}

We first introduce some notation, conventions and auxiliary results. Then we
set about to prove Theorems \ref{Theorem general} and \ref{Theorem main}.

\subsection{Notation, definitions and auxiliary results\label{subsection
notation}}

Throughout $H$ denotes a Hilbert space. The term \textit{norm} and the
notation $\left\Vert \cdot \right\Vert $ and $\left\langle \cdot,\cdot
\right\rangle$ always refer to the Euclidean norm and inner product on $%
\mathbb{R}^{K}$ or on $H$. Other norms are characterized by subscripts. If $%
H_{1}$ and $H_{2}$ are any Hilbert spaces $\mathcal{L}\left(
H_{1},H_{2}\right)$ denotes the vector space of bounded linear
transformations from $H_{1}$ to $H_{2}$. If $H_{1}=H_{2}$ we just write $%
\mathcal{L}\left( H_{1}\right) = \mathcal{L}\left( H_{1},H_{1}\right)$. With 
$\mathcal{U}\left(H_{1},H_{2}\right)$ we denote the set of isometries in $%
\mathcal{L}\left(H_{1},H_{2}\right)$, that is maps $U$ satisfying $%
\left\Vert Ux\right\Vert_{H_2}=\left\Vert x\right\Vert_{H_1}$ for all $x\in
H_{1}$.

We use $\mathcal{L}_{2}\left( H\right) $ for the set of Hilbert-Schmidt
operators on $H$, which becomes itself a Hilbert space with the inner
product $\left\langle T,S\right\rangle _{2}=$tr$\left( T^{\ast }S\right) $
and the corresponding (Frobenius) norm $\left\Vert \cdot \right\Vert _{2}$.

For $x\in H$ the rank-one operator $Q_{x}$ is defined by $%
Q_{x}z=\left\langle z,x\right\rangle x$. For any $T\in \mathcal{L}_{2}\left(
H\right) $ the identity 
\begin{equation*}
\left\langle T^{\ast }T,Q_{x}\right\rangle _{2}=\left\Vert Tx\right\Vert
^{2} 
\end{equation*}
is easily verified.

Suppose that $Y\subseteq \mathbb{R}^{K}$ spans $\mathbb{R}^{K}$.
%, that $% H^{\prime }$ is any Hilbert space (which could also be $\mathbb{R}^{K}$). 
It is easily verified that the quantity 
\begin{equation*}
\left\Vert T\right\Vert _{Y}=\sup_{y\in Y}\left\Vert Ty\right\Vert
\end{equation*}
defines a norm on $\mathcal{L}\left(\mathbb{R}^{K},H\right)$.

We use the following well known result on covering numbers (see, for
example, Proposition 5 in \cite{Cucker Smale 2001}).

\begin{proposition}
\label{Proposition covering}Let $B$ be a ball of radius $r$ in an $N$%
-dimensional Banach space and $\epsilon >0$. There exists a subset $%
B_{\epsilon }\subset B$ such that $\left\vert B_{\epsilon }\right\vert \leq
\left( 4r/\epsilon \right) ^{N}$ and $\forall z\in B,\exists z^{\prime }\in
B_{\epsilon }$ with $d(z,z^{\prime }) \leq \epsilon $, where $d$ is the
metric of the Banach space.
\end{proposition}

The following concentration inequality, known as the bounded difference
inequality \cite{McDiarmid 1998}, goes back to the work of Hoeffding \cite%
{Hoeffding 1963}.

\begin{theorem}
\label{Theorem Bded Difference}Let $\mu _{i}$ be a probability measure on a
space $\mathcal{X}_{i}$, for $i=1,\dots,m$. Let $\mathcal{X} =\prod_{i=1}^{m}%
\mathcal{X}_{i} $ and $\mu =\otimes _{i=1}^{m}\mu_{i}$ be the product space
and product measure respectively. Suppose the function $\Psi :\mathcal{X}
\rightarrow \mathbb{R}$ satisfies 
\begin{equation*}
\left\vert \Psi \left( \mathbf{x}\right) -\Psi \left( \mathbf{x}^{\prime
}\right) \right\vert \leq c_{i}
\end{equation*}
whenever $\mathbf{x}$ and $\mathbf{x}^{\prime } \in \mathcal{X}$ differ only in the $i$-th
coordinate, where $c_1,\dots,c_m$ are some positive parameters. Then 
\begin{equation*}
\Pr_{\mathbf{x}\sim \mu }\left\{ \Psi \left( \mathbf{x}\right) -\mathbb{E}_{ 
\mathbf{x}^{\prime }\sim \mu }\Psi \left( \mathbf{x}^{\prime }\right) \geq
t\right\} \leq \exp \left( \frac{-2t^{2}}{\sum_{i=1}^{m}c_{i}^{2}}\right) .
\end{equation*}
\end{theorem}

Throughout $\sigma _{i}$ will denote a sequence of mutually independent
random variables, uniformly distributed on $\left\{-1,1\right\} $ and $%
\gamma_{i}$, $\gamma _{ij}$ will be (multiple indexed) sequences of mutually
independent Gaussian random variables, with zero mean and unit standard
deviation.

If $\mathcal{F}$ is a class of real-valued functions on a space $\mathcal{X}$
and $\mu $ a probability measure on $\mathcal{X}$ then for $m\in \mathbb{N} $
the Rademacher and Gaussian complexities of $\mathcal{F}$ w.r.t. $\mu $ are
defined (\cite{Ledoux 1991},\cite{Bartlett 2002}) as 
\begin{eqnarray*}
\mathcal{R}_{m}\left( \mathcal{F},\mu \right) &=&\frac{2}{m}\mathbb{E}_{ 
\mathbf{x}\sim \mu ^{m}}\mathbb{E}_{\sigma }\sup_{f\in \mathcal{F}
}\sum_{i=1}^{m}\sigma _{i}f\left( x_{i}\right) \text{, } \\
\Gamma _{m}\left( \mathcal{F},\mu \right) &=&\frac{2}{m}\mathbb{E}_{\mathbf{%
\ x}\sim \mu ^{m}}\mathbb{E}_{\gamma }\sup_{f\in \mathcal{F}
}\sum_{i=1}^{m}\gamma _{i}f\left( x_{i}\right)
\end{eqnarray*}
respectively.

Appropriately scaled Gaussian complexities can be substituted for Rademacher
complexities, by virtue of the next Lemma. For a proof see, for example, 
\cite[p. 97]{Ledoux 1991}.

\begin{lemma}
\label{Lemma Gauss dominates Rademacher}For $Y\subseteq \mathbb{R}^{k}$ we
have $\mathcal{R}\left( Y\right) \leq \sqrt{\pi /2}~\Gamma \left( Y\right) $.
\end{lemma}

The next result is known as Slepian's lemma (\cite{Slepian}, \cite{Ledoux
1991}).

\begin{theorem}
\label{Slepian Lemma}Let $\Omega $ and $\Xi $ be mean zero, separable
Gaussian processes indexed by a common set $\mathcal{S}$, such that%
\begin{equation*}
\mathbb{E}\left( \Omega _{s_{1}}-\Omega _{s_{2}}\right) ^{2}\leq \mathbb{E}%
\left( \Xi _{s_{1}}-\Xi _{s_{2}}\right) ^{2}\text{ for all }s_{1},s_{2}\in 
\mathcal{S}\text{.}
\end{equation*}%
Then%
\begin{equation*}
\mathbb{E}\sup_{s\in \mathcal{S}}\Omega _{s}\leq \mathbb{E}\sup_{s\in 
\mathcal{S}}\Xi _{s}.
\end{equation*}
\end{theorem}

The following result, which generalizes Theorem 8 in \cite{Bartlett 2002},
plays a central role in our proof. 

\begin{theorem}
\label{Corollary Bound Finite Max Expectation} Let $\left\{ \mathcal{F}%
_{n}:1\leq n\leq N\right\} $ be a finite collection of $\left[ 0,b\right] $%
-valued function classes on a space $\mathcal{X}$, and $\mu $ a probability
measure on $\mathcal{X}$. Then $\forall \delta \in \left( 0,1\right) $ we
have with probability at least $1-\delta $ that%
\begin{equation*}
\max_{n\leq N}\sup_{f\in \mathcal{F}_{n}}\left[ \mathbb{E}_{x\sim \mu
}f\left( x\right) -\frac{1}{m}\sum_{i=1}^{m}f\left( x_{i}\right) \right]
\leq \max_{n\leq N}\mathcal{R}_{m}\left( \mathcal{F}_{n},\mu \right) +b\sqrt{%
\frac{\ln N+\ln \left( 1/\delta \right) }{2m}}.
\end{equation*}
\end{theorem}

\begin{proof}
Denote with $\Psi _{n}$ the function on $\mathcal{X}^{m}$ defined by%
\begin{equation*}
\Psi _{n}\left( \mathbf{x}\right) =\sup_{f\in \mathcal{F}_{n}}\left[ \mathbb{%
E}_{x\sim \mu }f\left( x\right) -\frac{1}{m}\sum_{i=1}^{m}f\left(
x_{i}\right) \right] ,\text{ }\mathbf{x}\in \mathcal{X}^{m}.
\end{equation*}%
By standard symmetrization (see, for example, \cite{van der Vaart 1996}) we
have $\mathbb{E}_{\mathbf{x}\sim \mu ^{m}}\Psi _{n}\left( \mathbf{x}\right)
\leq \mathcal{R}_{m}\left( \mathcal{F}_{n},\mu \right) \leq \max_{n\leq N}%
\mathcal{R}_{m}\left( \mathcal{F}_{n},\mu \right) $. Modifying one of the $%
x_{i}$ can change the value of any $\Psi _{n}\left( \mathbf{x}\right) $ by
at most $b/m$, so that by a union bound and the bounded difference
inequality (Theorem \ref{Theorem Bded Difference}) 
\begin{equation*}
\Pr \left\{ \max_{n\leq N}\Psi _{n}>\max_{n\leq N}\mathcal{R}_{m}\left( 
\mathcal{F}_{n},\mu \right) +t\right\} \leq \sum_{n}\Pr \left\{ \Psi _{n}>%
\mathbb{E}\Psi _{n}+t\right\} \leq Ne^{-2m\left( t/b\right) ^{2}}.
\end{equation*}%
Solving $\delta =Ne^{-2m\left( t/b\right) ^{2}}$ for $t$ gives the result.%
\qed
\end{proof}

Notice that replacing the functions $f\in \mathcal{F}_{n}$ by $b-f$ does not
affect the Rademacher complexities, so the above result can be used in a
two-sided way.

The following lemma was used in Section \ref{subsection nonnegative matrix
factorization}.

\begin{lemma}
\label{Little Lemma}Suppose $\left\Vert x\right\Vert \leq 1$, $\left\Vert
c_{k}\right\Vert =1,$ $\left\langle c_{k},c_{l}\right\rangle \geq 0$, $y\in 
\mathbb{R} ^{K}$, $y_{i}\geq 0$. If $y$ minimizes%
\begin{equation*}
h\left( y\right) =\left\Vert x-\sum_{k=1}^{K}y_{k}c_{k}\right\Vert ^{2},
\end{equation*}
then $\left\Vert y\right\Vert \leq 1$.
\end{lemma}

\begin{proof}
Assume that $y$ is a minimizer of $h$ and $\left\Vert y\right\Vert >1$.Then 
\begin{equation*}
\left\Vert \sum_{k=1}^{K}y_{k}c_{k}\right\Vert ^{2}=\left\Vert y\right\Vert
^{2}+\sum_{k\neq l}y_{k}y_{l}\left\langle c_{k},c_{l}\right\rangle >1.
\end{equation*}
Let the real-valued function $f$ be defined by $f\left( t\right) =h\left(
ty\right) $. Then 
\begin{eqnarray*}
f^{\prime }\left( 1\right) &=&2\left( \left\Vert
\sum_{k=1}^{K}y_{k}c_{k}\right\Vert ^{2}-\left\langle
x,\sum_{k=1}^{K}y_{k}c_{k}\right\rangle \right) \\
&\geq &2\left( \left\Vert \sum_{k=1}^{K}y_{k}c_{k}\right\Vert
^{2}-\left\Vert \sum_{k=1}^{K}y_{k}c_{k}\right\Vert \right) \\
&=&2\left( \left\Vert \sum_{k=1}^{K}y_{k}c_{k}\right\Vert -1\right)
\left\Vert \sum_{k=1}^{K}y_{k}c_{k}\right\Vert \\
&>&0\text{.}
\end{eqnarray*}
So $f$ cannot have a minimum at $1$, whence $y$ cannot be a minimizer of $h$%
. \qed
\end{proof}

\subsection{Proof of the main results}

We now fix a spanning codebook $Y\subseteq \mathbb{R}^{K}$ and recall that,
for $T\in \mathcal{L}\left( \mathbb{R}^{K},H\right) $, we had introduced the
notation 
\begin{equation*}
f_{T}\left( x\right) =\inf_{y\in Y}\left\Vert x-Ty\right\Vert ^{2},x\in H%
\text{.}
\end{equation*}%
Our principal object of study is the function class 
\begin{equation*}
\mathcal{F}=\left\{ f_{T}:T\in \mathcal{T}\right\} \text{,}
\end{equation*}%
where $\mathcal{T}\subset \mathcal{L}\left( \mathbb{R}^{K},H\right) $ is
some fixed set of candidate implementations of our coding scheme. We first
address the rather general Theorem \ref{Theorem general} which can be
treated in parallel to the case of $K$-means clustering. We begin with a
technical lemma.

\begin{lemma}
\label{Lemma Gaussian Complexity bounds}Suppose that

\begin{enumerate}
\item $\left( e_{k}: 1\leq k\leq K \right) $ is an orthonormal basis of $%
\mathbb{R}^{K}$;

\item $\mathcal{T}$ is the class of linear operators $T:\mathbb{R}%
^{K}\rightarrow H$ with $\left\Vert Te_{k}\right\Vert \leq c$;

\item $\left( x_{i}: 1 \leq i \leq m \right) $ is a sequence $x_{i}\in H$, $%
\left\Vert x_{i}\right\Vert \leq 1$;

\item $\left( \gamma _{ik}: 1 \leq i \leq m,~1\leq k\leq K \right) $ and $%
\left( \gamma _{ikl}:1\leq i\leq m,~1\leq k,l\leq K \right)$ are
orthogaussian sequences.
\end{enumerate}

Then the following three inequalities hold 
\begin{eqnarray*}
\mathbb{E}_{\gamma }\sup_{T\in \mathcal{T}}\sum_{i=1}^{m}\sum_{k=1}^{K}%
\gamma _{ik}\left\langle x_{i},Te_{k}\right\rangle &\leq &cK\sqrt{m} \\
\mathbb{E}_{\gamma }\sup_{T\in \mathcal{T}}\sum_{i=1}^{m}\sum_{k=1}^{K}%
\gamma _{ik}\left\Vert Te_{k}\right\Vert ^{2} &\leq &c^{2}K\sqrt{m} \\
\mathbb{E}_{\gamma }\sup_{T\in \mathcal{T}}\sum_{i=1}^{m}\sum_{k,l=1}^{K}%
\gamma _{ikl}\left\langle Te_{k},Te_{l}\right\rangle &\leq &c^{2}K^{2}\sqrt{m%
}.
\end{eqnarray*}
\end{lemma}

\begin{proof}
Using Cauchy-Schwarz' and Jensen's inequalities and the orthogaussian
properties of the $\gamma _{ik}$, we get%
\begin{equation*}
\mathbb{E}_{\gamma }\sup_{T\in \mathcal{T}}\sum_{k=1}^{K}\sum_{i=1}^{m}%
\gamma _{ik}\left\langle x_{i},Te_{k}\right\rangle \leq c\mathbb{E}_{\gamma
}\sum_{k=1}^K\left\Vert \sum_{i=1}^m\gamma _{ik}x_{i}\right\Vert \leq cK%
\sqrt{m}
\end{equation*}%
which is the first inequality. Similarly we obtain%
\begin{eqnarray*}
\mathbb{E}_{\gamma }\sup_{T\in \mathcal{T}}\sum_{k=1}^{K}\sum_{i=1}^{m}%
\gamma _{ik}\left\Vert Te_{k}\right\Vert ^{2} &\leq &c^{2}\mathbb{E}_{\gamma
}\sum_{k=1}^{K}\left\vert \sum_{i=1}^{m}\gamma _{ik}\right\vert \leq c^{2}K%
\sqrt{m} \\
\mathbb{E}_{\gamma }\sup_{T\in \mathcal{T}}\sum_{k,l=1}^{K}\sum_{i=1}^{m}%
\gamma _{ikl}\left\langle Te_{k},Te_{l}\right\rangle &\leq &c^{2}\mathbb{E}%
_{\gamma }\sum_{k,l=1}^{K}\left\vert \sum_{i=1}^{m}\gamma _{ikl}\right\vert
\leq c^{2}K^{2}\sqrt{m}.
\end{eqnarray*}
\qed
\end{proof}

\begin{proposition}
\label{Proposition general complexities}Suppose that the probability measure 
$\mu $ is supported on the unit ball of $H$, that $\left\{ e_{k}: 1 \leq k
\leq K\right\} $ is an orthonormal basis of $\mathbb{R}^{K}$ and that $%
\mathcal{T}$ is a class of linear operators $T:\mathbb{R}^{K}\rightarrow H$
with $\left\Vert Te_{k}\right\Vert \leq c$ for $1 \leq k \leq K$, with $%
c\geq 1$. Let $Y$ be a nonempty closed subset of the unit ball in $\mathbb{R}%
^{K}$ and 
\begin{equation*}
\mathcal{F}_{Y}=\left\{ x\in H\mapsto \min_{y\in Y}\left\Vert
x-Ty\right\Vert ^{2}:T\in \mathcal{T}\right\} .
\end{equation*}%
Then 
\begin{equation*}
\mathcal{R}\left( \mathcal{F}_{Y},\mu \right) \leq 6c^{2}K^{2}\sqrt{\frac{%
\pi }{m}}.
\end{equation*}%
and if $Y=\left\{ e_{k}: 1 \leq k \leq K\right\} $ then the bound improves to%
\begin{equation*}
\mathcal{R}\left( \mathcal{F}_{Y},\mu \right) \leq c^{2}K\sqrt{\frac{18\pi }{%
m}}.
\end{equation*}
\end{proposition}

\begin{proof}
By Lemma \ref{Lemma Gauss dominates Rademacher} it suffices to bound the
corresponding Gaussian averages, which we shall do using Slepian's Lemma
(Theorem \ref{Slepian Lemma}). First fix a sample $\mathbf{x}$ and define
Gaussian processes $\Omega $ and $\Xi $ indexed by $\mathcal{T}$
\begin{eqnarray*}
\Omega _{T} &=&\sum_{i}\gamma _{i}\min_{y}\left\Vert x_{i}-Ty\right\Vert ^{2}%
\text{ and} \\
\Xi _{T} &=&\sqrt{8}\sum_{ik}\gamma _{ik}\left\langle
x_{i},Te_{k}\right\rangle +\sqrt{2}\sum_{ilk}\gamma _{ilk}\left\langle
Te_{l},Te_{k}\right\rangle .
\end{eqnarray*}%
Suppose $T_{1},T_{2}\in \mathcal{T}$. For any $x\in H$ we have, using $%
\left( a+b\right) ^{2}\leq 2a^{2}+2b^{2}$ and Cauchy-Schwarz 
\begin{eqnarray*}
&&\left( \min_{y\in Y}\left\Vert x-T_{1}y\right\Vert ^{2}-\min_{y}\left\Vert
x-T_{2}y\right\Vert ^{2}\right) ^{2} \\
&\leq &\left( \max_{y\in Y}\left\Vert x-T_{1}y\right\Vert ^{2}-\left\Vert
x-T_{2}y\right\Vert ^{2}\right) ^{2} \\
&\leq &8\max_{y\in Y}\left( \sum_{k}y_{k}\left\langle x,\left(
T_{1}-T_{2}\right) e_{k}\right\rangle \right) ^{2}+2\max_{y\in Y}\left(
\sum_{kl}y_{k}y_{l}\left\langle e_{k},\left( T_{1}^{\ast }T_{1}-T_{2}^{\ast
}T_{2}\right) e_{l}\right\rangle \right) ^{2} \\
&\leq &8\sum_{k}\left( \left\langle x,T_{1}e_{k}\right\rangle -\left\langle
x,T_{2}e_{k}\right\rangle \right) ^{2}+2\sum_{kl}\left( \left\langle
T_{1}e_{k},T_{1}e_{l}\right\rangle -\left\langle
T_{2}e_{k},T_{2}e_{l}\right\rangle \right) ^{2}.
\end{eqnarray*}%
We therefore have%
\begin{eqnarray*}
\mathbb{E}\left( \Omega _{T_{1}}-\Omega _{T_{2}}\right) ^{2}
&=&\sum_{i}\left( \min_{y}\left\Vert x_{i}-T_{1}y\right\Vert
^{2}-\min_{y}\left\Vert x_{i}-T_{2}y\right\Vert ^{2}\right) ^{2} \\
&\leq &8\sum_{ik}\left( \left\langle x_{i},T_{1}e_{k}\right\rangle
-\left\langle x_{i},T_{2}e_{k}\right\rangle \right) ^{2}+2\sum_{ikl}\left(
\left\langle T_{1}e_{k},T_{1}e_{l}\right\rangle -\left\langle
T_{2}e_{k},T_{2}e_{l}\right\rangle \right) ^{2} \\
&=&\mathbb{E}\left( \Xi _{T_{1}}-\Xi _{T_{2}}\right) ^{2}.
\end{eqnarray*}%
So, by Slepian's Lemma and the first and last inequalities in Lemma \ref%
{Lemma Gaussian Complexity bounds}%
\begin{eqnarray*}
\mathbb{E}\sup_{T\in \mathcal{T}}\Omega _{T} &\leq &\mathbb{E}\sup_{T\in 
\mathcal{T}}\Xi _{T} \\
&\leq &\sqrt{8}\mathbb{E}\sup_{T\in \mathcal{T}}\sum_{ik}\gamma
_{ik}\left\langle x_{i},Te_{k}\right\rangle +\sqrt{2}\mathbb{E}\sup_{T\in 
\mathcal{T}}\sum_{ilk}\gamma _{ilk}\left\langle Te_{l},Te_{k}\right\rangle \\
&\leq &cK\sqrt{8m}+c^{2}K^{2}\sqrt{2m}.
\end{eqnarray*}%
Multiply by $\sqrt{2\pi }/m$ to get a bound on the Rademacher complexity of 
\begin{equation*}
\mathcal{R}\left( \mathcal{F}_{Y},\mu \right) \leq 4cK\sqrt{\frac{\pi }{m}}%
+2c^{2}K^{2}\sqrt{\frac{\pi }{m}}\leq 6c^{2}K^{2}\sqrt{\frac{\pi }{m}}.
\end{equation*}%
To obtain the second conclusion we improve the bound on the Gaussian
average. With $\Omega _{T}$ as above we set 
\begin{equation*}
\Xi _{T}=\sum_{i=1}^{m}\sum_{k=1}^{K}\gamma _{ik}\left\Vert
x_{i}-Te_{k}\right\Vert ^{2}\text{.}
\end{equation*}
Now we have for $T_{1},T_{2}\in \mathcal{T}$ that 
\begin{eqnarray*}
\mathbb{E}\left(\Omega_{T_{1}}-\Omega _{T_{2}}\right)^{2} &=&
\sum_{i=1}^{m}\left( \min_{k=1}^K \Vert x_{i}-T_{1}e_{k}\Vert
^{2}-\min_{k=1}^K\Vert x_{i}-T_{2}e_{k}\Vert ^{2}\right) ^{2} \\
&\leq &\sum_{i=1}^{m}\max_{k=1}^{K}\left( \left\Vert
x_{i}-T_{1}e_{k}\right\Vert ^{2}-\left\Vert x_{i}-T_{2}e_{k}\right\Vert
^{2}\right) ^{2} \\
&\leq &\sum_{i=1}^{m}\sum_{k=1}^{K}\left( \left\Vert
x_{i}-T_{1}e_{k}\right\Vert ^{2}-\left\Vert x_{i}-T_{2}e_{k}\right\Vert
^{2}\right) ^{2} \\
&=&\mathbb{E}\left( \Xi _{T_{1}}-\Xi _{T_{2}}\right) ^{2}\text{.}
\end{eqnarray*}%
Again with Slepian's Lemma and the triangle inequality 
\begin{align*}
\mathbb{E}_{\gamma }\sup_{T\in \mathcal{T}}\Omega _{T}& \leq \mathbb{E}%
_{\gamma }\sup_{T\in \mathcal{T}}\Xi _{T}=\mathbb{E}_{\gamma }\sup_{T\in 
\mathcal{T}}\sum_{i=1}^{m}\sum_{k=1}^{K}\gamma _{ik}\left\Vert
x_{i}-Te_{k}\right\Vert ^{2} \\
& \leq 2\mathbb{E}_{\gamma }\sup_{T\in \mathcal{T}}\sum_{i=1}^{m}%
\sum_{k=1}^{K}\gamma _{ik}\left\langle x_{i},Te_{k}\right\rangle +\mathbb{E}%
_{\gamma }\sup_{T\in \mathcal{T}}\sum_{i=1}^{m}\sum_{k=1}^{K}\gamma
_{ik}\left\Vert Te_{k}\right\Vert ^{2} \\
& \leq 3c^{2}K\sqrt{m},
\end{align*}%
where the last inequality follows from the first two inequalities in Lemma %
\ref{Lemma Gaussian Complexity bounds}. Multiply by $\sqrt{2\pi }/m$ as above%
\qed
\end{proof}

Theorem \ref{Theorem general} follows from observing that the functions in $%
\mathcal{F}$ map to $\left[ 0,4c^{2}\right] $ and combining the above bound
on the Rademacher complexity with Theorem \ref{Corollary Bound Finite Max
Expectation} with $N=1$ and $b=4$.

The second conclusion of the proposition yields a bound for $K$-means
clustering, corresponding to the choices $Y=\left\{ e_{1},\dots
,e_{K}\right\} $ and $\mathcal{T}=\left\{ T:\left\Vert Te_{k}\right\Vert
\leq 1,~1\leq k\leq K\right\} $. As already noted in Section \ref{subsection
kmeans clustering} the vectors $Te_{k}$ define the cluster centers. With
Theorem \ref{Corollary Bound Finite Max Expectation} we obtain

\begin{theorem}
\label{Theorem Kmeans clustering}For every $\delta >0$ with probability
greater $1-\delta $ in the sample $\mathbf{x}\sim \mu ^{m}$ we have for all $%
T\in \mathcal{T}$ 
\begin{equation*}
\mathbb{E}_{x\sim \mu }\min_{k=1}^{K}\left\Vert x-Te_{k}\right\Vert ^{2}\leq 
\frac{1}{m}\sum_{i=1}^{m}\min_{k=1}^{K}\left\Vert x_{i}-Te_{k}\right\Vert
^{2}+K\sqrt{\frac{18\pi }{m}}+\sqrt{\frac{8\ln \left( 1/\delta \right) }{m}}.
\end{equation*}%
\bigskip
\end{theorem}

To prove Theorem \ref{Theorem main} a more subtle approach is necessary. The
idea is the following: every implementing map $T\in \mathcal{T}$ can be
factored as $T=US$, where $S$ is a $K\times K$ matrix, $S\in \mathcal{L}%
\left( \mathbb{R}^{K}\right) $, and $U$ is an isometry, $U\in {\mathcal{U}}(%
\mathbb{R}^{K},H)$. Suitably bounded $K\times K$ matrices form a compact,
finite dimensional set, the complexity of which can be controlled using
covering numbers, while the complexity arising from the set of isometries
can be controlled with Rademacher and Gaussian averages. Theorem \ref%
{Corollary Bound Finite Max Expectation} then combines these complexity
estimates.

For fixed $S\in \mathcal{L}\left(\mathbb{R}^{K}\right) $ we denote 
\begin{equation*}
\mathcal{G}_{S}=\left\{ f_{US}:U\in \mathcal{U}\left(\mathbb{R}^{K},H\right)
\right\} .
\end{equation*}
Recall the notation $\left\Vert \mathcal{T}\right\Vert _{Y}=\sup_{T\in 
\mathcal{T}}\left\Vert T\right\Vert _{Y}=\sup_{T\in \mathcal{T}}\sup_{y\in
Y}\left\Vert Ty\right\Vert $. With $\mathcal{S}$ we denote the set of $%
K\times K$ matrices 
\begin{equation*}
\mathcal{S}=\left\{ S\in \mathcal{L}\left( \mathbb{R}^{K}\right) :\left\Vert
S\right\Vert _{Y}\leq \left\Vert \mathcal{T} \right\Vert _{Y}\right\} \text{.%
}
\end{equation*}

\begin{lemma}
\label{Lemma Key} Assume $\left\Vert \mathcal{T}\right\Vert _{Y}\geq 1$,
that the functions in $\mathcal{F}$, when restricted to the unit ball of $H$%
, have range contained in $\left[ 0,b\right] $, and that the measure $\mu $
is supported on the unit ball of $H$. Then with probability at least $%
1-\delta $ we have for all $T\in \mathcal{T}$ that 
\begin{multline*}
\mathbb{E}_{x\sim \mu }f_{T}\left( x\right) -\frac{1}{m}\sum_{i=1}^{m}f_{T}
\left( x_{i}\right) \\
\leq \sup_{S\in \mathcal{S}}\mathcal{R}_{m}\left( \mathcal{G}_{S},\mu
\right) +\frac{bK}{2}\sqrt{\frac{\ln \left( 16m\left\Vert \mathcal{T}
\right\Vert _{Y}^{2}\right) }{m}}+\frac{8\left\Vert \mathcal{T}\right\Vert
_{Y}}{\sqrt{m}}+b\sqrt{\frac{\ln \left( 1/\delta \right) }{2m}}.
\end{multline*}
\end{lemma}

\begin{proof}
Fix $\epsilon >0$. The set $\mathcal{S}$ is the ball of radius $\left\Vert 
\mathcal{T}\right\Vert _{Y}$ in the $K^{2}$-dimensional Banach space $\left( 
\mathcal{L}\left(\mathbb{R}^{K}\right) ,\left\Vert .\right\Vert _{Y}\right) $
so by Proposition \ref{Proposition covering} we can find a subset $\mathcal{S%
}_{\epsilon }\subset \mathcal{S}$, of cardinality $\left\vert \mathcal{S}%
_{\epsilon }\right\vert \leq \left( 4\left\Vert \mathcal{T}\right\Vert
_{Y}/\epsilon \right) ^{K^{2}} $ such that every member of $\mathcal{S}$ can
be approximated by a member of $\mathcal{S}_{\epsilon }$ up to distance $%
\epsilon $ in the norm $\left\Vert .\right\Vert _{Y}$.

We claim that for all $T\in \mathcal{T}$ there exist $U\in {\mathcal{U}}( 
\mathbb{R}^{K},H)$ and $S_{\epsilon }\in \mathcal{S}_{\epsilon }$ such that 
\begin{equation*}
\left\vert f_{T}\left( x\right) -f_{US_{\epsilon }}\left( x\right)
\right\vert <4\left\Vert \mathcal{T}\right\Vert _{Y}\epsilon,
\end{equation*}
for all $x$ in the unit ball of $H$. To see this write $T=US$ with $U\in {\ 
\mathcal{U}} (\mathbb{R}^{K},H)$ and $S\in \mathcal{L}(\mathbb{R}^{K})$.
Then, since $U$ is an isometry, we have 
\begin{equation*}
\left\Vert S\right\Vert _{Y}=\sup_{y\in Y}\left\Vert Sy\right\Vert
=\sup_{y\in Y}\left\Vert Ty\right\Vert =\left\Vert T\right\Vert _{Y}\leq
\left\Vert \mathcal{T}\right\Vert _{Y}
\end{equation*}
so that $S\in \mathcal{S}$. We can therefore choose $S_{\epsilon }\in 
\mathcal{S}_{\epsilon }$ such that $\left\Vert S_{\epsilon }-S\right\Vert
_{Y}<\epsilon $. Then for $x\in H$, with $\left\Vert x\right\Vert \leq 1$,
we have 
\begin{eqnarray*}
\left| f_{T}\left( x\right) -f_{US_{\epsilon }}\left( x\right) \right|
&=&\left| \inf_{y\in Y}\left( \left\Vert x-USy\right\Vert^{2} \right)
-\inf_{y\in Y}\left( \left\Vert x-US_{\epsilon }y\right\Vert^{2} \right)
\right| \\
&\leq &\sup_{y\in Y}\left|\left( \left\Vert x-USy\right\Vert ^{2}-\left\Vert
x-US_{\epsilon }y\right\Vert ^{2}\right)\right| \\
&=&\sup_{y\in Y}\left|\left\langle US_{\epsilon }y-USy,2x-\left( USy+US_{\epsilon
}y\right) \right\rangle \right|\\
&\leq &\left( 2+2\left\Vert \mathcal{T}\right\Vert _{Y}\right) \sup_{y\in
Y}\left\Vert \left( S_{\epsilon }-S\right) y\right\Vert \leq 4\left\Vert 
\mathcal{T}\right\Vert _{Y}\epsilon .
\end{eqnarray*}
Apply Theorem \ref{Corollary Bound Finite Max Expectation} to the finite
collection of function classes $\left\{ \mathcal{G}_{S}:S\in \mathcal{S}
_{\epsilon }\right\} $ to see that with probability at least $1-\delta $ 
\begin{eqnarray*}
&&\sup_{T\in \mathcal{T}}\mathbb{E}_{x\sim \mu }f_{T}\left( x\right) -\frac{%
1 }{m}\sum_{i=1}^{m}f_{T}\left( x_{i}\right) \\
&\leq &\max_{S\in \mathcal{S}_{\epsilon }}\sup_{U\in \mathcal{U}\left( 
\mathbb{R} ^{K},H\right) }\mathbb{E}_{x\sim \mu }f_{US}\left( x\right) -%
\frac{1}{m} \sum_{i=1}^{m}f_{US}\left( x_{i}\right) +8\left\Vert \mathcal{T}%
\right\Vert _{Y}\epsilon \\
&\leq &\max_{S\in \mathcal{S}_{\epsilon }}\mathcal{R}_{m}\left( \mathcal{G}
_{S},\mu \right) +b\sqrt{\frac{\ln \left\vert \mathcal{S}_{\epsilon
}\right\vert +\ln \left( 1/\delta \right) }{2m}}+8\left\Vert \mathcal{T}
\right\Vert _{Y}\epsilon \\
&\leq &\sup_{S\in \mathcal{S}}\mathcal{R}_{m}\left( \mathcal{G}_{S},\mu
\right) +\frac{bK}{2}\sqrt{\frac{\ln \left( 16m\left\Vert \mathcal{T}
\right\Vert _{Y}^{2}\right) }{m}}+\frac{8\left\Vert \mathcal{T}\right\Vert
_{Y}}{\sqrt{m}}+b\sqrt{\frac{\ln \left( 1/\delta \right) }{2m}},
\end{eqnarray*}
where the last line follows from the known bound on $\left\vert \mathcal{S}
_{\epsilon }\right\vert $, subadditivity of the square root and the choice $%
\epsilon =1/\sqrt{m}$.\qed\bigskip
\end{proof}

\begin{remark}
If $H$ is finite dimensional the above result may be improved to 
\begin{equation}
\mathbb{E} f_{T} - {\hat{\mathbb{E}}} f_{T} \leq \frac{b}{2}\sqrt{\frac{dK
\ln \left( 16m\left\Vert \mathcal{T} \right\Vert _{Y}^{2}\right) }{m}}+\frac{%
8\left\Vert \mathcal{T}\right\Vert _{Y}}{\sqrt{m}}+b\sqrt{\frac{\ln \left(
1/\delta \right) }{2m}}.  \label{eq:FD}
\end{equation}
To see this, follow the same lines as in Lemma \ref{Lemma Key} to note that 
\begin{equation*}
\sup_{T\in \mathcal{T}} \mathbb{E} f_T - {\hat{\mathbb{E}}}f_T \leq \max_{T
\in \mathcal{T}_\epsilon} \mathbb{E} f_T - {\hat{\mathbb{E}}}f_T + 8 \|%
\mathcal{T}\|_Y \epsilon,
\end{equation*}
where $\mathcal{T}_{\epsilon }$ is a subset of $\mathcal{T}$ such that every
member of $\mathcal{T}$ can be approximated by a member of $\mathcal{T}%
_{\epsilon }$ up to distance $\epsilon $ in the norm $\left\Vert\cdot\right%
\Vert_{Y}$.

By Proposition \ref{Proposition covering}, $\left\vert \mathcal{T}_{\epsilon
}\right\vert \leq \left( 4\left\Vert \mathcal{T}\right\Vert _{Y}/\epsilon
\right)^{dK}$. Inequality \eqref{eq:FD} now follows from Theorem \ref%
{Corollary Bound Finite Max Expectation} with $N=|\mathcal{T}_\epsilon|$ and 
$\epsilon =1/\sqrt{m}$.
\end{remark}

To complete the proof of Theorem \ref{Theorem main} we now fix some $S\in 
\mathcal{S}$ and focus on the corresponding function class $\mathcal{G}_{S}$.

\begin{lemma}
For any $S\in \mathcal{L}\left(\mathbb{R}^{K}\right) $ we have 
\begin{equation*}
\mathcal{R}\left( \mathcal{G}_{S},\mu \right) \leq 2 \sqrt{2\pi} \left\Vert
S\right\Vert _{Y} \frac{K}{\sqrt{m}}.
\end{equation*}
\end{lemma}

\begin{proof}
Let $\left\Vert x_{i}\right\Vert \leq 1$ and define Gaussian processes $%
\Omega _{U}$ and $\Xi _{U}$ indexed by ${\mathcal{U}}(\mathbb{R}^{K},H)$ 
\begin{eqnarray*}
\Omega _{U} &=&\sum_{i=1}^{m}\gamma _{i}\inf_{y\in Y}\left\Vert
x_{i}-USy\right\Vert ^{2} \\
\Xi _{U} &=&2\left\Vert S\right\Vert _{Y}\sum_{k=1}^{K}\sum_{i=1}^{m}\gamma
_{ik}\left\langle x_{i},Ue_{k}\right\rangle \text{,}
\end{eqnarray*}%
where the $e_{k}$ are the canonical basis of $\mathbb{R}^{K}$. For $%
U_{1},U_{2}\in {\mathcal{U}}({\mathbb{R}}^{K},H)$ we have 
\begin{eqnarray*}
\mathbb{E}\left( \Omega _{U_{1}}-\Omega _{U_{2}}\right) ^{2} &\leq
&\sum_{i=1}^{m}\left( \sup_{y\in Y}\Vert x_{i}-U_{1}Sy\Vert ^{2}-\Vert
x_{i}-U_{2}S\Vert ^{2}\right) ^{2} \\
&\leq &\sum_{i=1}^{m}\sup_{y\in Y}4\langle x_{i},(U_{2}-U_{1})Sy\rangle ^{2}
\\
&\leq &4\sum_{i=1}^{m}\sup_{y\in Y}\Vert U_{2}^{\ast }x_{i}-U_{1}^{\ast
}x_{i}\Vert ^{2}\Vert Sy\Vert ^{2} \\
&=&4\left\Vert S\right\Vert _{Y}^{2}\sum_{i=1}^{m}\sum_{k=1}^{K}\left(
\left\langle x_{i},U_{1}e_{k}\right\rangle -\left\langle
x_{i},U_{2}e_{k}\right\rangle \right) ^{2} \\
&=&\mathbb{E}\left( \Xi _{U_{1}}-\Xi _{U_{2}}\right) ^{2}.
\end{eqnarray*}%
It follows from Lemma \ref{Lemma Gauss dominates Rademacher} and Slepians
lemma (Theorem \ref{Slepian Lemma}) that 
\begin{equation*}
\mathcal{R}_{m}\left( \mathcal{G}_{S},\mu \right) \leq \mathbb{E}_{\mathbf{x}%
\sim \mu ^{m}}\frac{2}{m}\sqrt{\frac{\pi }{2}}\mathbb{E}_{\mathbf{\gamma }%
}\sup_{U}\Xi _{U},
\end{equation*}%
so the result follows from the following inequalities, using Cauchy-Schwarz'
and Jensen's inequality, the orthonormality of the $\gamma _{ik}$ and the
fact that $\left\Vert x_{i}\right\Vert \leq 1$ on the support of $\mu $. 
\begin{eqnarray*}
\mathbb{E}_{\mathbf{\gamma }}\sup_{U}\Xi _{U} &=&2\left\Vert S\right\Vert
_{Y}\mathbb{E}\sup_{U}\sum_{k=1}^{K}\left\langle \sum_{i=1}^{m}\gamma
_{ik}x_{i},Ue_{k}\right\rangle \\
&\leq &2\left\Vert S\right\Vert _{Y}\sum_{k=1}^{K}\mathbb{E}\left\Vert
\sum_{i=1}^{m}\gamma _{ik}x_{i}\right\Vert \\
&\leq &2\left\Vert S\right\Vert _{Y}K\sqrt{m}.
\end{eqnarray*}%
\qed
\end{proof}

Substitution of the last result in Lemma \ref{Lemma Key} and noting that,
for $K\geq 1$, $2\sqrt{2\pi }K+8\leq 14K$, gives Theorem \ref{Theorem main}.

Observe that when the set $\mathcal{S}$ contains only the identity matrix,
the function class $\mathcal{G}_S$ is the class of reconstruction errors of
PCA. In this case, the result can be improved as shown by the next lemma.

\begin{lemma}
\label{Lemma PCA}$\mathcal{R}\left( \mathcal{D},\mu \right) \leq 2\sqrt{K/m}$%
.
\end{lemma}

\begin{proof}
Recall, for every $z\in H$, that the outer product operator $Q_{z}$ is
defined by $Q_{z}x=\left\langle x,z\right\rangle z$. With $\left\langle
\cdot,\cdot\right\rangle _{2}$ and $\left\Vert \cdot \right\Vert _{2}$
denoting the Hilbert-Schmidt inner product and norm respectively we have for 
$\left\Vert x_{i}\right\Vert \leq1$ 
\begin{eqnarray*}
\mathbb{E}_{\sigma }\sup_{f\in \mathcal{D}}\sum_{i=1}^{m}\sigma _{i}f\left(
x_{i}\right) &=&\mathbb{E}_{\sigma }\sup_{U\in \mathcal{U}
}\sum_{i=1}^{m}\sigma _{i}\left( \left\Vert x_{i}\right\Vert ^{2}-\left\Vert
UU^{\ast }x_{i}\right\Vert ^{2}\right) \\
&=&\mathbb{E}_{\sigma }\sup_{U\in \mathcal{U}}\left\langle
\sum_{i=1}^{m}\sigma _{i}Q_{x_{i}},UU^{\ast }\right\rangle _{2} \\
&\leq &\mathbb{E}_{\sigma }\left\Vert \sum_{i=1}^{m}\sigma
_{i}Q_{x_{i}}\right\Vert _{2}\sup_{U\in \mathcal{U}}\left\Vert UU^{\ast
}\right\Vert _{2} \\
&\leq &\sqrt{mK},
\end{eqnarray*}
since the Hilbert-Schmidt norm of a $K$-dimensional projection is $\sqrt{K}$%
. The result follows upon multiplication with $2/m$ and taking the
expectation in $\mu ^{m}$.\qed
\end{proof}

An application of Theorem \ref{Corollary Bound Finite Max Expectation} with $%
N=1$ and $b=1$ also give a generalization bound for PCA of order $\sqrt{K/m}$%
.

\section{Concluding remarks}

We have analyzed a general method to encode random vectors in a Hilbert
space $H$. The method searches for an operator $T:\mathbb{R}^{K}\rightarrow H
$ which minimizes, within some prescribed class $\mathcal{T}$, the empirical
average of the reconstruction error, which is defined as the minimum
distance between a given point in $H$ and an image of the operator $T$
acting on a prescribed codebook $Y$.

We have presented two approaches to upper bound the estimation error of the
method in terms of the parameter $K$, the sample size $m$ and the properties
of the sets $\mathcal{T}$ and $Y$. The first approach is based on a direct
bound for the Rademacher average of the loss class induced by the
reconstruction error. The bound matches the best known bound for $K$-means
clustering in a Hilbert space \cite{Biau Devroye Logosi 2006} but also
applies to other interesting coding techniques such as sparse coding and
non-negative matrix factorization. The second approach uses a decomposition
of the function class as a union of function classes parameterized by 
$K$-dimensional isometries. The main idea is to approximate the union with a
finite union via covering numbers and then bound the complexity of each
class under the union with Rademacher averages. This second result is more
complicated than the first one, however it provides in certain cases a better
dependency of the bound on the parameter $K$ at the expense of an additional logarithmic
factor in $m$.

We conclude with some open problems and possible extensions which are
suggested by this study. Firstly, it would be valuable to investigate the
possibility of removing the logarithmic term in $m$ in the bound of Theorem 
\ref{Theorem main}. Secondly, it would be important to elucidate whether the
dependency in $K$ in the same bound is optimal. The latter problem is
also mentioned in \cite{Biau Devroye Logosi 2006} in the case of
$K$-means clustering. Finally, in would be interesting to study
possible improvements of our results in the case that additional
assumptions on the probability measure $\mu$ are introduced. For
example, in the case of $K$-means clustering in a finite dimensional
Hilbert space \cite{antos2005} shows that for certain classes of
probability measures the rate of convergence can be improved to
$O(\log(m)/m)$ and it may be possible to obtain similar improvements
in our general framework.

\subsection*{Acknowledgments}

This work was supported by EPSRC Grants GR/T18707/01 and EP/D071542/1.

\end{document}